\documentclass[conference]{IEEEtran}
\usepackage{cite}
\usepackage{amsmath,amssymb,amsfonts}
\usepackage{algorithmic}
\usepackage{graphicx}
\usepackage{textcomp}
\usepackage{xcolor}
\usepackage{authblk}
\usepackage{subcaption}
\usepackage{hyperref}

\def\BibTeX{{\rm B\kern-.05em{\sc i\kern-.025em b}\kern-.08em
    T\kern-.1667em\lower.7ex\hbox{E}\kern-.125emX}}

\title{Improving Driver Drowsiness Detection via Personalized EAR/MAR Thresholds and CNN-Based Classification
}

\author{
{
Gökdeniz Ersoy, Mehmet Alper Tatar, Eray Tonbul, Serap Kırbız\textsuperscript{*}} \\
\\
MEF University \\
Electrical and Electronics Engineering Department \\
Huzur Mah, Maslak Ayazağa Cd. No:4, 34396 Sarıyer/İstanbul \\
\href{mailto:ersoyg@mef.edu.tr} {ersoyg@mef.edu.tr}, \href{mailto:tatarme@mef.edu.tr}{tatarme@mef.edu.tr}, \href{mailto:tonbule@mef.edu.tr}{tonbule@mef.edu.tr}, \href{mailto:kirbizs@mef.edu.tr}{kirbizs@mef.edu.tr}\\
\textsuperscript{*}ORCID: \href{https://orcid.org/0000-0001-7718-3683}{0000-0001-7718-3683}}

\begin{document}
\maketitle

\begin{abstract}
Driver drowsiness is a major cause of traffic accidents worldwide, posing a serious threat to public safety. Vision-based driver monitoring systems often rely on fixed Eye Aspect Ratio (EAR) and Mouth Aspect Ratio (MAR) thresholds; however, such fixed values frequently fail to generalize across individuals due to variations in facial structure, illumination, and driving conditions. This paper proposes a personalized driver drowsiness detection system that monitors eyelid movements, head position, and yawning behavior in real time and provides warnings when signs of fatigue are detected. The system employs driver-specific EAR and MAR thresholds, calibrated before driving, to improve classical metric-based detection. In addition, deep learning–based Convolutional Neural Network (CNN) models are integrated to enhance accuracy in challenging scenarios. The system is evaluated using publicly available datasets as well as a custom dataset collected under diverse lighting conditions, head poses, and user characteristics. Experimental results show that personalized thresholding improves detection accuracy by 2–3\% compared to fixed thresholds, while CNN-based classification achieves 99.1\% accuracy for eye state detection and 98.8\% for yawning detection, demonstrating the effectiveness of combining classical metrics with deep learning for robust real-time driver monitoring.

\end{abstract}

\begin{IEEEkeywords}
Fatigue detection, Sleep detection, drowsiness detection, EAR, PERCLOS, MAR, Facial Landmark, CNN
\end{IEEEkeywords}

\section{Introduction}
\label{sec:intro}

According to the World Health Organization (WHO), an average of 3260 people die every day from traffic accidents. Studies indicate that approximately 10\% to 30\% of these accidents occur as a result of driver drowsiness or falling asleep while driving \cite{tekin2022effect}. 

Vision-based driver monitoring systems have gained considerable attention due to their non-intrusive nature and low hardware requirements. Facial cues such as eye closure, yawning, and head position are commonly used indicators of driver fatigue. Classical methods based on Eye Aspect Ratio (EAR), Mouth Aspect Ratio (MAR), and PERCLOS enable efficient real-time analysis using facial landmark geometry \cite{cech2016real}. However, most existing approaches rely on fixed threshold values, which often fail to generalize across individuals due to differences in facial structure, eye shape, mouth geometry, and illumination conditions. 

To overcome these limitations,personalized thresholding strategies have been proposed, in which EAR and MAR values are adapted to individual facial characteristics. In parallel, deep learning–based approaches, particularly Convolutional Neural Networks (CNNs) \cite{gomaa2022cnn, kirbiz2025}, have been widely adopted for eye and mouth state classification, offering improved robustness under challenging conditions. Despite their effectiveness, these two approaches are often studied independently, and their comparative performance in real-time driver monitoring scenarios has not been sufficiently analyzed.

In this paper, we investigate two vision-based approaches for driver drowsiness detection: (i) a personalized facial landmark–based method using EAR, MAR, and PERCLOS and (ii) a CNN-based classification approach for eye and yawning detection. Both methods are implemented and evaluated within a unified experimental framework. The personalized approach introduces a short calibration phase to determine user-specific EAR and MAR thresholds, while the CNN-based approach relies on learned visual features extracted from eye and mouth images.

The main contributions of this paper can be summarized as follows:

\begin{itemize}
    \item A personalized driver drowsiness detection method based on facial landmarks that adapts to individual facial characteristics.
    \item A CNN-based eye and yawning classification approach for robust fatigue detection.
    \item A comparative experimental evaluation of personalized threshold-based and CNN-based methods under real-time driving conditions.
\end{itemize}

The remainder of this paper is organized as follows. Section \ref{sec:background} reviews related work on driver drowsiness detection. Section \ref{sec:proposed}  describes the proposed personalized hybrid framework. Section \ref{sec:test} presents the experimental setup and datasets, while Section \ref{sec:result} discusses the results. Section \ref{sec:conclusion} concludes the paper and outlines future research directions.

\section{BACKGROUND INFORMATION}
\label{sec:background}
Several methods have been proposed for driver fatigue and drowsiness detection. These methods can be categorized into vehicle-based  and vision-based methods \cite{albadawi2022review}. Vehicle-based methods \cite{baccour2022comparative} monitor parameters such as the lane tracking, steering wheel contact, pedal control, etc. However, their performance is highly dependent on road and vehicle conditions. 

\begin{figure}
 \centering
 \includegraphics[width=\columnwidth]{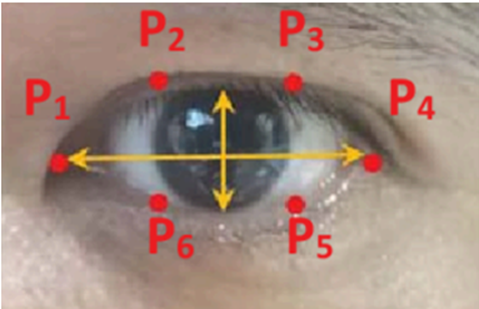}
 \caption{The facial landmark points around the eye.}
 \label{fig:eye_landmark}
\end{figure}

Vision-based fatigue detection systems commonly rely on facial cues such as eye closure, yawning and head position \cite{makhmudov2024real}. Among these, the Eye Aspect Ratio (EAR) and Mouth Aspect Ratio (MAR) are widely used geometric features derived from facial landmark points. EAR, introduced by Soukupová and Čech \cite{cech2016real}, is an effective method for determining whether the eyes are open or closed using six landmark points around the eyelid. It is defined as
\begin{equation}
    EAR=\frac{||p_2-p_6||+||p_3-p_5||}{2||p_1-p_4||},
    \label{eq:EAR}
\end{equation}
where the landmark points are illustrated in Figure \ref{fig:eye_landmark}. EAR measures the ratio between the vertical and horizontal distances of the eye. When the eyes are open, the EAR value remains relatively high; when the eyes close, the ratio decreases significantly. In practical systems, a threshold is used to distinguish between open and closed eye states. A continuous decrease below this threshold indicates prolonged eye closure, which is associated with fatigue or drowsiness rather than normal blinking.

To further quantify eye closure behavior, the Percentage of Eye Closure (PERCLOS) metric is commonly employed \cite{abe2023perclos}. PERCLOS reflects slow eyelid closures rather than brief blinks and has been shown to be a reliable indicator of driver fatigue. It is defined as

\begin{equation}
    PERCLOS=\frac{T_c}{T_T}\times 100,
    \label{eq:perclos}
\end{equation}
where $T_c$ is the total time the eyes are closed and $T_T$ is the total monitoring duration.

\begin{figure}
\centering
\includegraphics[width=\columnwidth]{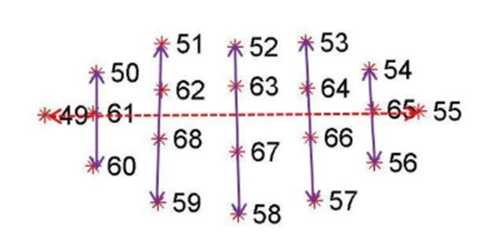}
\caption{The facial landmark points around the mouth used for MAR calculation.}
\label{fig:mouth_landmark}
\end{figure}

Yawning behavior is another important indicator of driver drowsiness and is commonly detected using the Mouth Aspect Ratio (MAR) \cite{omidyeganeh2016yawning}. MAR is calculated from facial landmark points located around the mouth and is defined as

\begin{equation}
MAR=\frac{||p_{63}-p_{67}||+||p_{64}-p_{66}||}{2||p_{61}-p_{65}||},
\label{eq:MAR}
\end{equation}
where the corresponding landmark points are illustrated in Fig.~\ref{fig:mouth_landmark}. When a driver yawns, the vertical mouth opening increases significantly, resulting in a higher MAR value \cite{omidyeganeh2016yawning}. Similar to EAR-based detection, MAR-based systems typically rely on threshold values to distinguish yawning from normal mouth movements.

A major limitation of EAR and MAR based approaches is their reliance on fixed threshold values, which often fail to generalize across different individuals due to variations in facial geometry and illumination conditions. To address this issue, several studies have  proposed personalized or adaptive thresholding strategies, where EAR and MAR thresholds are adjusted according to individual facial characteristics, improving robustness and reducing false detections \cite{xu2025detecting}.

In parallel with geometric approaches, deep learning–based methods have been widely adopted for driver drowsiness detection. Convolutional Neural Networks (CNNs) have demonstrated high accuracy in classifying eye and mouth images as open/closed or yawning/normal \cite{gomaa2022cnn}. Unlike landmark-based methods, CNNs learn discriminative features directly from image data and are generally more robust to noise and illumination changes. However, CNN-based approaches typically require higher computational resources and larger training datasets.

In this study, personalized threshold–based methods and CNN-based classification are treated as distinct approaches for driver drowsiness detection. Their performance is analyzed and compared within a unified experimental framework to evaluate their respective strengths and limitations in real-time driving scenarios.

\section{PROPOSED METHODS}
\label{sec:proposed}
This section describes the two vision-based approaches considered for driver drowsiness detection: (i) a personalized facial landmark–based method using EAR, MAR, and PERCLOS metrics, and (ii) a CNN-based eye and yawning classification method. Both approaches are implemented and evaluated under identical experimental conditions.

\subsection{Personalized Facial Landmark–Based Drowsiness Detection}
\label{sec:personalized}
The proposed personalized method extracts 468 facial landmarks using MediaPipe Face Mesh \cite{jakhete2024comprehensive}. The facial landmarks around the eyes are used to calculate the EAR as in Eq. (\ref{eq:EAR}). The facial landmarks around the lip are used to calculate the MAR using Eq. (\ref{eq:MAR}). 

To account for individual differences in facial structure, a short personalized calibration phase is conducted before monitoring. During the calibration phase, baseline EAR and MAR values are recorded while the driver is seated in the vehicle with a neutral facial expression prior to continuous driving. Personalized thresholds are then computed based on these baselines. Eye closure is detected when EAR falls below the personalized threshold for a sustained period, and prolonged eye closure is quantified using the PERCLOS metric. Yawning is detected when MAR exceeds the personalized threshold. Head position is also monitored using facial landmark geometry to detect downward head tilts associated with fatigue. If the vertical distance between the tip of the nose and eye level exceeds a certain limit, a head down detected warning is triggered. 

This approach improves robustness across different users and lighting conditions while maintaining low computational requirements suitable for real-time applications.

\subsection{CNN-Based Eye and Yawning Classification}
\label{sec:CNN}
\begin{figure}
 \centering
 \includegraphics[width=\columnwidth]{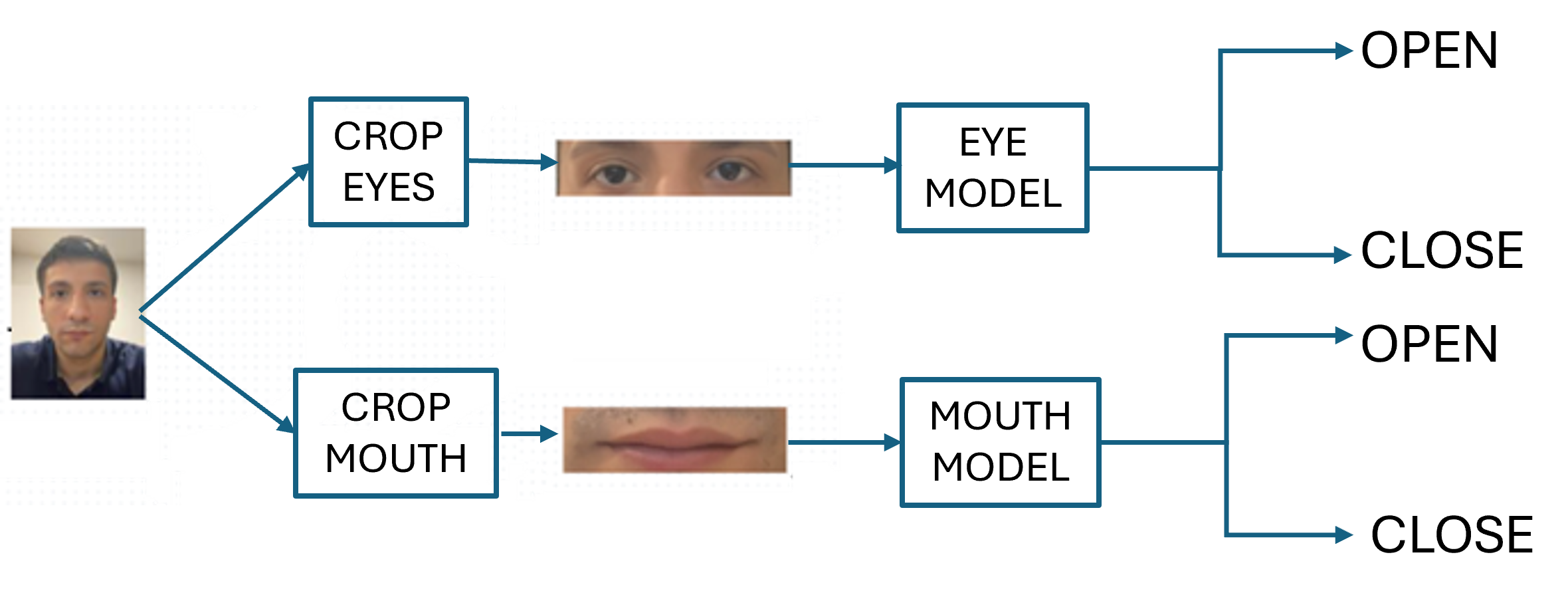}
 \caption{The flowchart of the proposed system}
 \label{fig:proposed}
\end{figure}

The CNN-based approach operates independently from the personalized threshold method. First, eye and mouth regions are cropped based on the detected facial landmarks and resized to a fixed resolution suitable for CNN input as seen in Figure \ref{fig:proposed}. Two separate Convolutional Neural Networks (CNNs) are trained:

\begin{itemize}
\item Eye CNN: Classifies eye state as open or closed.
\item Mouth CNN: Classifies mouth state as yawning or not yawning.
\end{itemize}

Unlike geometric thresholding, the CNN models learn visual features directly from images, providing enhanced robustness to illumination changes, partial occlusions, and variations in facial appearance. Drowsiness is inferred by analyzing the temporal patterns of the classified states, such as repeated eye closures or frequent yawning events.

\subsection{Decision Logic and Warning Generation}
\label{sec:decision}
For both approaches, temporal smoothing is applied to reduce sensitivity to transient events such as blinks or brief mouth openings. When blinking or talking, the eyes close or mouth opens. To avoid warnings in such sudden changes, a buffer of 15 frames is used and average values are calculated. When sustained signs of drowsiness (eye close, mouth open or head nodding) are detected, the system triggers auditory and visual warnings to alert the driver. The warning mechanism is implemented identically for both approaches to allow fair comparison during experimental evaluation.

\section{EXPERIMENTS}
\label{sec:test}
This section describes the datasets, experimental setup and evaluation methodology used to assess the performance of the proposed personalized and CNN-based driver drowsiness detection methods.
\subsection{Datasets}
\label{sec:dataset}
The system is evaluated using both publicly available datasets and a custom dataset collected under realistic driving conditions:
\begin{enumerate}
\item {\bf MRL Eye Dataset} \cite{mrl_eye_dataset} is a large-scale dataset containing 84,898 eye images in open and closed states under varying illumination conditions.
It is widely used for eye detection and blink analysis. 
\item {\bf Yawn Dataset} \cite{yawn_dataset}  contains 5,119 mouth images labeled as open or closed. It is used for yawning detection. 
\item {\bf Custom Dataset} was collected using a frontal camera in a vehicle simulator and real driving scenarios. The dataset includes approximately 1,000 images from multiple participants with different facial structures, lighting conditions, and head poses. Each image is labeled for eye state and yawning state.
\end{enumerate}

All datasets were split into 70\% training, 15\% validation, and 15\% testing sets. Before model training, facial landmarks were detected using MediaPipe Face Mesh \cite{jakhete2024comprehensive}, and the eye and mouth regions were subsequently cropped for feature extraction or input to the CNN model. Additionally, Contrast Limited Adaptive Histogram Equalization (CLAHE) \cite{sharma2023review} was applied to enhance image contrast.

\subsection{Experimental Setup}
\label{sec:setup}
For personalized threshold–based detection, each participant undergoes a short calibration phase for 5 seconds seated in a neutral posture with eyes open and mouth closed. The system calculates the average EAR and MAR values specific to the driver. The personalized thresholds are determined as 75 \% of the EAR value and 140 \% of the MAR value based on multiple trials. Eye closure is monitored via EAR and PERCLOS, yawning via MAR, and abnormal head positions are detected using landmark-based head pose estimation. If the detected EAR and MAR values exceed these thresholds, or when a downward head tilt, associated with fatigue, drowsiness is detected. 

For CNN-based detection, the eye and mouth regions are cropped and resized to $64 \times 64$ pixels. Two separate CNNs are trained: one for eye state and one for yawning. Data augmentation techniques \cite{kayaouglu2023cnn} such as horizontal flipping, random brightness adjustment, and small rotations are applied to improve model generalization.

All experiments are conducted in real-time simulation and recorded video settings to mimic practical driving conditions. Both methods generate continuous outputs, which are post-processed using temporal smoothing to reduce false positives from transient events (e.g., blinks or brief mouth openings).
\subsection{Evaluation Metrics}
\label{sec:metric}
The performance of both methods is evaluated using classification accuracy. Accuracy is defined as
\begin{equation}
Accuracy=\frac{TP+TN}{TP+TN+FP+FN},
\label{eq:qccuracy}
\end{equation}
where TP, TN, FP and FN corresponds to True Positives, True Negatives, False Positives and False Negatives, respectively.

In addition to accuracy, confusion matrices are provided for deeper analysis between open and closed eye states, as well as yawning detection for the general and personalized EAR and MAR based models.

All methods are tested under identical conditions to assess performance and real-time applicability.

\subsection{Comparison Strategy}
To ensure a fair comparison:
\begin{enumerate}
\item Same participants and dataset splits are used for both methods.
\item  Temporal smoothing and warning logic are identical across methods.
\item  Both methods are tested under varying lighting conditions, head poses, and facial structures to evaluate robustness.
\item  Key performance differences, such as improved accuracy for CNN under challenging conditions or reduced false alarms for personalized thresholding, are analyzed in Section \ref{sec:result}.
\end{enumerate}

\section{Results and Discussion}
\label{sec:result}

This section presents a comparison of the personalized threshold–based and CNN-based driver drowsiness detection methods. The evaluation uses public and custom datasets (Section~\ref{sec:dataset}) and focuses on classification accuracy for eye closure and yawning detection.

\begin{figure}[h]
    \centering
    \begin{subfigure}[t]{0.99\columnwidth}
        \centering
        \includegraphics[width=\linewidth]{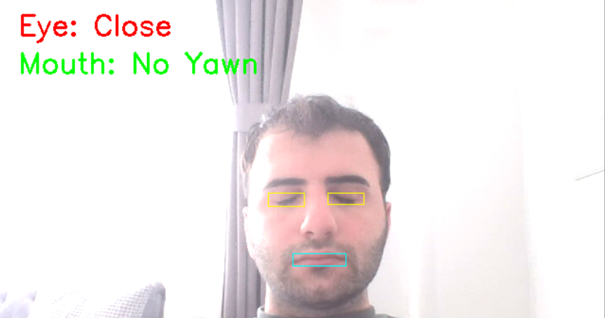}
        \caption{Detection of eye closure.}
        \label{fig:eye_closed}
    \end{subfigure}
    \hfill
    \begin{subfigure}[t]{0.99\columnwidth}
        \centering
        \includegraphics[width=\linewidth]{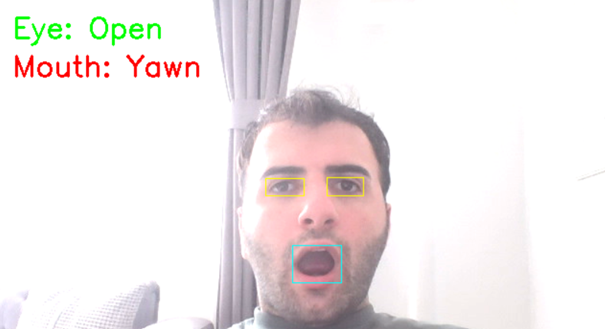}
        \caption{Detection of yawning.}
        \label{fig:mouth_open}
    \end{subfigure}
    \caption{Examples of real-time detection outputs for (a) eye closure and (b) yawning.}
    \label{fig:example}
\end{figure}

Figure~\ref{fig:example} shows real-time detection outputs. The system successfully identifies (a) eye closure and (b) yawning events. Temporal smoothing and a short buffer prevent false warnings caused by brief blinks or talking.

Table~\ref{tab:eye_performance} summarizes the accuracy of the EAR based method for generalized threshold, EAR based method for personalized threshold, and CNN-based methods for eye state detection. Personalized thresholding improves accuracy compared to generalized thresholds by around 1.53 \% , while the CNN model improves accuracy by 7.40 \%, achieving the highest performance.  

\begin{table}[h]
\centering
\caption{Performance comparison of eye state detection methods.}
\label{tab:eye_performance}
\begin{tabular}{lcc}
\hline
\textbf{Method} & \textbf{Approach} & \textbf{Accuracy (\%)}  \\
\hline
Generalized EAR & Threshold-based & 91.70\\
Personalized EAR & Threshold-based & 93.23  \\
CNN (Eye) & Learning-based & 99.10  \\
\hline
\end{tabular}
\end{table}
\begin{table}[h!]
\centering
\caption{Performance comparison of yawning detection methods. }
\label{tab:mouth_performance}
\begin{tabular}{lccc}
\hline
\textbf{Method} & \textbf{Approach} & \textbf{Accuracy (\%)} \\
\hline
Generalized MAR      & Threshold-based & 95.90 \\
Personalized MAR     & Threshold-based & 97.23 \\
CNN (Mouth)          & Learning-based  & 98.80 \\
\hline
\end{tabular}
\end{table}

Similarly, Table~\ref{tab:mouth_performance} presents the results for yawning detection based on generalized MAR threshold, personalized MAR threshold and CNN mouth model. Personalized MAR thresholds enhance detection accuracy relative to generalized thresholds, while CNN-based models achieve the highest overall performance.

\begin{figure}[h!]
    \centering
    \includegraphics[width=0.5\textwidth]{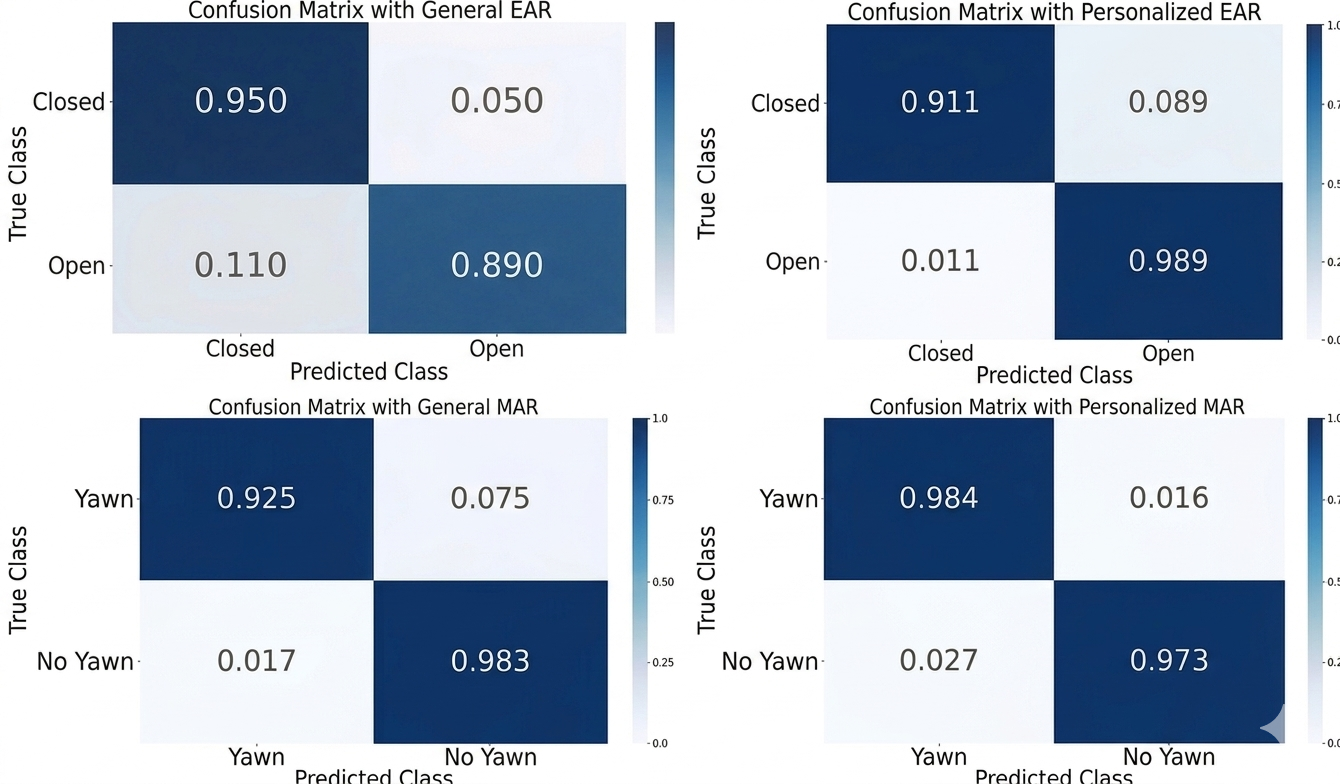}
    \caption{Confusion matrices comparing general and personalized models for EAR and MAR metrics.}
    \label{fig:confusion}
\end{figure}
The confusion matrices in Figure \ref{fig:confusion} further support these findings. For EAR-based detection, the generalized model achieves 95.0\% accuracy for closed eyes and 89.0\% for open eyes, while the personalized model improves open-eye accuracy to 98.9\%.

For MAR-based detection, the general model reaches 92.5\% accuracy in yawn detection, which increases to 98.4\% with personalization. High accuracy is also preserved for the ‘No Yawn’ class (98.3\% and 97.3\%), indicating that personalization improves detection performance while maintaining overall accuracy.

\subsection{Discussion}

The results highlight the following points:

\begin{itemize}
    \item Accuracy: CNN-based models provide the highest accuracy for both eye and yawning detection, demonstrating robustness to variations in lighting, head pose, and facial geometry.
    \item Efficiency: Personalized thresholding improves accuracy over generalized thresholds while requiring minimal computational resources, making it suitable for real-time applications.
    \item Real-time applicability: Temporal smoothing effectively prevents false positives due to blinks, talking, or brief mouth movements.
    \item Complementary strengths: Combining classical metrics (EAR, MAR, PERCLOS) with CNN classification offers a trade-off between accuracy and computational cost.
\end{itemize}

Overall, these experiments confirm that driver-specific calibration reduces classification errors across both metrics (EAR and MAR) and improves classical metric-based detection, while CNN-based methods ensure high accuracy and robustness in challenging conditions.

\section{Conclusion}
\label{sec:conclusion}
In this paper, we presented and evaluated two vision-based approaches for driver drowsiness detection: (i) a personalized facial landmark–based method using EAR, MAR, and PERCLOS metrics, and (ii) a CNN-based method for eye state and yawning classification. Both approaches were tested on public and custom datasets under varying illumination, head pose, and facial characteristics.

Experimental results show that the CNN-based method achieves the highest accuracy and robustness to environmental and individual variations. The personalized threshold-based method, while slightly less accurate, offers a lightweight and computationally efficient alternative suitable for real-time and low-resource applications. Personalized calibration improves detection reliability by adapting to individual differences.

These findings indicate that the two approaches provide complementary trade-offs between accuracy and computational cost, making them suitable for different deployment scenarios. Real-time implementation with temporal smoothing and warning mechanisms further demonstrates their practical applicability.

Future work will focus on improving performance under extreme conditions and extending the system with multimodal inputs, such as infrared imaging and physiological signals, as well as adaptive and online learning strategies.

Overall, the proposed framework provides an effective and adaptable solution for real-time driver drowsiness detection.

\bibliographystyle{IEEEtran}
\bibliography{refs}
\end{document}